\documentclass[10pt,twocolumn,letterpaper]{article}

\usepackage{iccv}              

\definecolor{iccvblue}{rgb}{0.21,0.49,0.74}
\usepackage[pagebackref,breaklinks,colorlinks,allcolors=iccvblue]{hyperref}

\def\paperID{3561} 
\def\confName{ICCV}
\def\confYear{2025}

\usepackage[accsupp]{axessibility}  

\usepackage{subcaption}
\usepackage{enumitem}
\usepackage{multirow} 
\usepackage{colortbl}  
\usepackage{soul} 
\usepackage{booktabs}
\usepackage{xcolor}
\usepackage{amsmath}
\usepackage{bbding} 
\usepackage{subcaption}
\usepackage{algorithm}
\usepackage{algpseudocode}
\usepackage{listings}

\definecolor{Lavender}{RGB}{148, 198, 205}

\colorlet{tableheadcolor}{gray!10} 
\colorlet{tablerowcolor}{Lavender!20} 
\newcommand{\headcol}{\rowcolor{tableheadcolor}}
\newcommand{\rowcol}{\rowcolor{tablerowcolor}}

\newcommand{\topline}{\arrayrulecolor{black}\specialrule{1pt}{\abovetopsep}{0pt}%
    \arrayrulecolor{tableheadcolor}\specialrule{\belowrulesep}{0pt}{0pt}%
    \arrayrulecolor{black}}

\title{Aligning Information Capacity Between Vision and Language via Dense-to-Sparse Feature Distillation for Image-Text Matching}

\author{Yang Liu\textsuperscript{\rm 1, \rm 2}, Wentao Feng\textsuperscript{\rm 1, \rm 2}, Zhuoyao Liu\textsuperscript{\rm 1, \rm 2}, Shudong Huang\textsuperscript{\rm 1, \rm 2, \thanks{Corresponding Author}}, Jiancheng Lv\textsuperscript{\rm 1, \rm 2}\\
\tt\small \textsuperscript{\rm 1}College of Computer Science, Sichuan University, China\\
\tt\small \textsuperscript{\rm 2}Engineering Research Center of Machine Learning and Industry Intelligence, China\\
\tt\small \url{https://d2s-vse.github.io}
\and
}

\begin{document}
\maketitle
\begin{abstract}
Enabling Visual Semantic Models to effectively handle multi-view description matching has been a longstanding challenge. 
Existing methods typically learn a set of embeddings to find the optimal match for each view's text and compute similarity. However, the visual and text embeddings learned through these approaches have limited information capacity and are prone to interference from locally similar negative samples.
To address this issue, we argue that the information capacity of embeddings is crucial and propose Dense-to-Sparse Feature Distilled Visual Semantic Embedding (D2S-VSE), which enhances the information capacity of sparse text by leveraging dense text distillation.
Specifically, D2S-VSE is a two-stage framework. In the pre-training stage, we align images with dense text to enhance the information capacity of visual semantic embeddings.
In the fine-tuning stage, we optimize two tasks simultaneously, distilling dense text embeddings to sparse text embeddings while aligning images and sparse texts, enhancing the information capacity of sparse text embeddings.
Our proposed D2S-VSE model is extensively evaluated on the large-scale MS-COCO and Flickr30K
datasets, demonstrating its superiority over recent state-of-the-art methods.
\end{abstract}    
\section{Introduction}
\label{sec:intro}

In the domain of vision and language, the alignment of vision and language stands as a foundational task, with the core objective of establishing a robust and precise bridge between these two distinct yet interconnected modalities. 
The traditional approach to cross-modal retrieval \cite{liu2022regularizing,  gu2018look, qin2022deep, qin2023cross} follows a standard procedure: encoding images and texts into a shared embedding space and measuring their similarity. Typically, models are optimized using a triplet loss, which ensures that matched image-text pairs exhibit higher similarity than unmatched pairs \cite{faghri2017vse++, kiros2014unifying}.

\begin{figure}
    \centering
    \includegraphics[width=\linewidth]{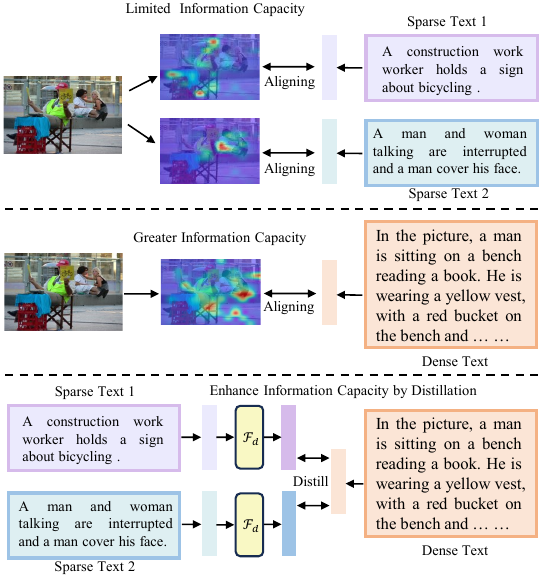}
    \caption{(Top) In previous methods, when addressing problems with varying information density, a set of visual embeddings is learned to match the most similar text embeddings. However, this approach results in learned embeddings with limited information capacity.
    (Mid) Since dense text has a higher information density than sparse text, training with aligned dense text results in image embeddings with greater information capacity.
    (Bottom) Sparse text embedding can be distilled through dense text embedding to enhance its information capacity, enabling the matching of text descriptions from different perspectives with a single visual embedding. (Best viewed in color)
}
    \label{fig:intro}
\end{figure}

However, most traditional Visual Semantic Embedding (VSE) models overlook the inherent differences between image and text modalities \cite{chen2021learning, radford2021learning, li2019visual}. Specifically, visual signals represent natural, objective records of real-world phenomena, whereas textual signals are inherently semantic and subjective, shaped by human interpretation. Consequently, the same image may be described in various ways depending on individual perspectives.  
Existing studies refer to this issue as the ``ambiguity issue" \cite{song2019polysemous, chun2023improved}, ``sparse annotations" \cite{kim2023improving} and ``multi-view descriptions" \cite{li2022multi, qu2020context}. But, at its core, the problem stems from the inherent differences between image and text modalities, particularly the disparity in ``information density" \cite{liu2025asymmetric}.


Recent studies \cite{chun2023improved, song2019polysemous, li2022multi,liu2025asymmetric} propose learning a sets of embeddings from images/texts, where each embedding captures specific semantics. For each image-text pair, the most relevant embedding is selected for similarity calculation. While this approach reduces computational overhead and partially addresses information density challenges, we identify a critical limitation: 
As shown in the figure \ref{fig:intro}, the text in the dataset is usually short and concise. The image and text embeddings learned in this way can only capture a limited subset of features, resulting in limited information capacity being learned, and are therefore easily disturbed by negative samples with similar local semantics.
Given this, we argue that the \textit{information capacity} of visual semantic embeddings is crucial for the image-text matching task. The key question, then, is: \textit{how can we learn visual semantic embeddings with higher information capacity?}

In this paper, we introduce a novel approach, Dense-to-Sparse Feature Distillation for Visual Semantic Embedding (D2S-VSE), which enhances visual semantic embeddings by aligning both semantics and information capacity.
Our key hypothesis posits that text embeddings derived from dense image captions inherently possess greater information capacity compared than those from sparse textual annotations in traditional datasets such as Flickr30K and COCO.
Specifically, we introduce a two-stage training framework.
In the pre-train stage, we generate dense descriptions for images using LLaVa and pre-train image-dense description pairs to align images with dense textual descriptions in the embedding space. This alignment enables the model to extract more comprehensive features, resulting in embeddings with higher information capacity.
In the fine-tuning stage, we introduce a dense-to-sparse feature distillation method. During this phase, while maintaining alignment between images and sparse text annotations from the dataset, we additionally distill knowledge from dense text embeddings into sparse text embeddings to enhance their information capacity.
We formulate this distillation process as a masked signal reconstruction task for sparse text embeddings. To achieve this, we implement a Transformer-based decoder module that follows the text encoder. This decoder processes short text features combined with learnable {mask} tokens, aiming to predict the latent context of the sparse text and thus enhance its information capacity.
And in the inference phase, we only require the image and the sparse text from the dataset, without the need for dense text, thus incurring no additional computational overhead.

Our main contributions are summarized as follows:
\begin{itemize}
    \item[$\bullet$] 
    \textbf{Importance of Information Capacity:} We identify the critical role of enhancing the information capacity of visual semantic embeddings. Based on this insight, we propose a novel framework to improve the VSE model by increasing the information capacity of visual semantic embeddings.
    \item[$\bullet$] 
    \textbf{Dense-to-Sparse Feature Distilled Visual Semantic Embedding (D2S-VSE): }We introduce D2S-VSE, a two-stage training approach that leverages dense text to enhance the information capacity of image embeddings in the first pre-training stage. In the second stage, a distillation branch is incorporated to improve the information capacity of sparse text embedding by distilling dense text embedding from sparse text embedding, ensuring both semantic alignment and balanced information capacity between image and text embeddings.
    \item[$\bullet$] \textbf{State-of-the-Art Performance:} We evaluate D2S-VSE with existing fine-grained methods across various model backbones. D2S-VSE outperforms the latest state-of-the-art methods on image-text retrieval benchmarks, including Flickr30K and MS-COCO.
\end{itemize}



\begin{figure*}[!t]
\begin{center}
\includegraphics[width=\linewidth]{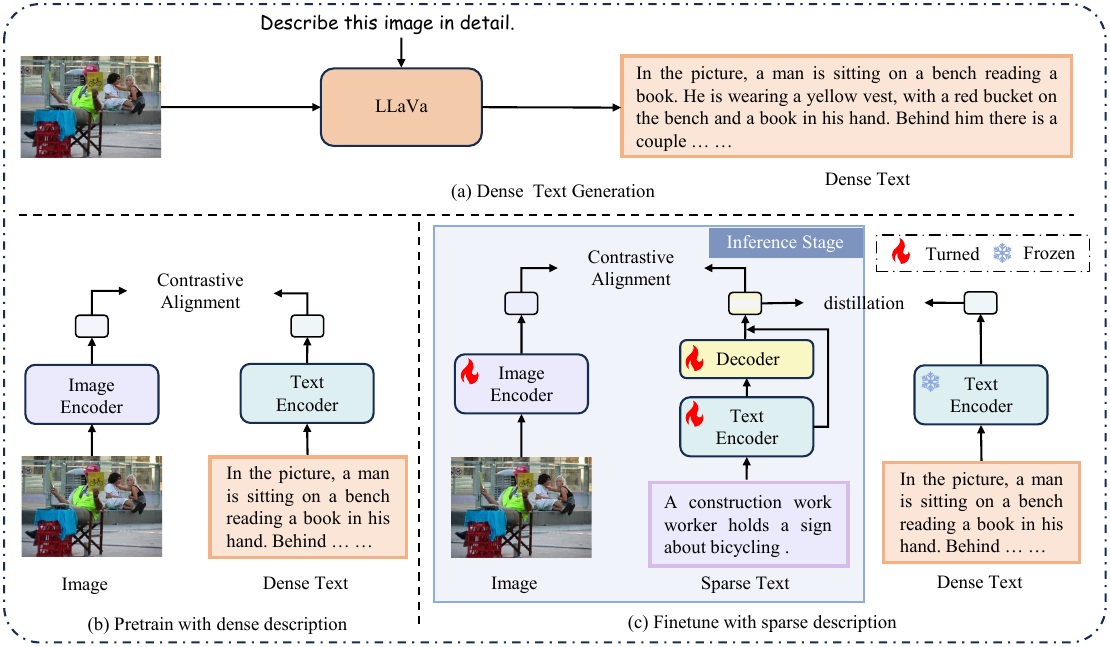}
\end{center}
\caption{Overview of our D2S-VSE framework. (a) \textit{Dense Text Generation:} D2S-VSE utilizes the LLaVa \cite{liu2023visual} to generate dense text descriptions for images in the dataset that contain all the details of the image. (b) \textit{Pre-train With Dense Descriptions:} D2S-VSE first uses contrastive learning to pre-train the model on image-dense-text pairs to enhance the information capacity of visual semantic embeddings. (c) \textit{Finetune With the Sparse Descriptions:} 
D2S-VSE performs distillation from dense to sparse while contrasting the alignment, allowing sparse text embeddings to learn more details of dense text embeddings, thereby improving the information capacity of sparse text embeddings and better aligning them with image embeddings.
}
\label{framework}
\end{figure*}

\section{Related Work}
\label{sec:related}
\subsection{Image-Text Matching}
Existing Image-text retrieval approaches can be broadly categorized into two groups based on the granularity of the matching similarity: \textit{Global-level matching} and \textit{Local-level matching} methods. Global-level matching methods project images and texts into a shared embedding space, computing similarity using improved loss functions \cite{faghri2017vse++,chun2021probabilistic} to bring semantically aligned pairs closer together. To improve the quality of the embedding space, recent related works proposed several more sophisticated networks, such as graph convolutional network \cite{liu2020graph, wang2020consensus}, Generalized Pooling Operator \cite{chen2021learning}, and other model architectures \cite{huang2018learning,li2019visual,li2022image}.
In order to facilitate the closer alignment of semantically corresponding image-text pairs. 
Different from global matching methods, local matching methods\cite{diao2021similarity,liu2023bcan,wang2019camp,wei2020universal,zhang2022negative,zhang2020context,chen2020imram,qu2021dynamic} focus more on the alignment between regions. Most of them capture the finer-grained relationship between images and texts by adopting cross-modal networks.
\\

\noindent \textbf{Information Density Issues.}
Recent studies have proposed various methods to address the issue of differing information density between images and text. PCME \cite{chun2021probabilistic} proposed a probabilistic cross-modal embedding to properly represent the one-to-many relationships in joint embedding. DivE \cite{kim2023improving} introduced a slot attention-based set prediction module, where the elements of the embedding set compete iteratively to aggregate the input data. Under the framework of PCME, PCME++ \cite{chun2023improved} introduced a new probabilistic distance with a closed-form solution. AVSE \cite{liu2025asymmetric} proposed a matching module for compute the dynamic similarity between asymmetric multi-view image and text embeddings. 

These methods learn a set of embeddedings to match ambiguous samples, but reduce the information capacity of each embedding.

\subsection{Vision language model with dense text.}
In the field of vision-language research, few studies have leveraged long-text descriptions to enhance the multimodal representation of images. DreamLip utilizes a multimodal large language model to generate dense text descriptions and randomly extracts sentence fragments for pre-training, but it does not fully exploit long-text pre-training. Long-CLIP fine-tunes CLIP—pre-trained on a short-text-image dataset—on long-text-image pairs to unlock long-text capabilities, yet it overlooks short-text effectiveness. LoTLIP directly pre-trains on long-text-image data and introduces corner tokens to aggregate diverse textual information, preserving the ability to handle short text.

Different from the above methods that focus on pre-training, our method utilizes a very small amount of long text for pre-training, aiming to use the long text to enhance the information capacity of visual semantic embeddings.


\section{Dense-to-Sparse Feature Distilled Visual Semantic Embedding}
Our goal is to enhance Visual Semantic Embedding (VSE) models to address the challenge of inconsistent information density between images and text. Existing VSE models typically align images with the sparse textual descriptions provided in datasets. This approach biases the image encoder toward textual perspectives, limiting its ability to fully comprehend visual content and making it difficult to handle ambiguous retrieval cases.
To overcome this issue, we propose a two-stage training procedure. An overview of our D2S-VSE framework is illustrated in Figure \ref{framework}.

\begin{itemize}
    \item[$\bullet$] \textbf{Stage \uppercase\expandafter{\romannumeral1}.} We leverage the pre-trained multimodal large model LLaVa to generate dense textual descriptions for images. The VSE model is then pre-trained using these \textbf{image-dense text pairs}. By training on data with comparable information density, the model learns embeddings with \textit{greater information capacity}, enabling a more comprehensive understanding of visual content (Sec. \ref{sec:3.1}).
    \item[$\bullet$] \textbf{Stage \uppercase\expandafter{\romannumeral2}.} We fine-tune the pre-trained VSE model on general datasets containing \textbf{image-sparse text pairs}. Simultaneously, we distill knowledge from the dense text features into the sparse text representations, enhancing their information capacity. While aligning image-text embeddings, we also align the \textit{information capacity} between visual and textual modalities. This enables the VSE model to effectively handle varying information densities in images and texts (Sec. \ref{sec:3.2}).
\end{itemize}

\subsection{Pre-training with Dense Descriptions}
\label{sec:3.1}
To address the issue of inconsistent information density between images and text, we introduce a pre-training stage where the Visual Semantic Embedding (VSE) model learns from image-dense text pairs. This stage enhances the model’s ability to extract rich semantic representations from images by ensuring that textual descriptions contain a level of detail comparable to the visual information.

\noindent \textbf{Dense Text Generation.}
To generate dense textual descriptions for images, we leverage the pre-trained multimodal large model LLaVa \cite{liu2023visual}, which integrates vision and language understanding to generate detailed descriptions based on image inputs. Unlike traditional dataset texts, which are often brief and sparse, LLaVa produces detailed descriptions that include object attributes, relationships, actions, and contextual information. 
Given a raw image $I$, we feed it into the LLaVa with the prompt \texttt{"Describe this image in detail"}. LLaVa then produces a dense textual description that captures fine-grained details of the image. The hyperparameters used for LLaVa are detailed in Sec \ref{details}.
This process ensures that the textual input provides a comprehensive representation of the image, reducing the information density gap between modalities.

\noindent \textbf{Image-Dense-Caption Pre-train.}
Once the dense captions are generated, we pre-train the VSE model using these image-dense text pairs. The objective of this stage is to enable the model to learn embeddings that effectively capture fine-grained visual semantics and their corresponding textual descriptions.

We employ a contrastive learning framework where the image encoder $\mathcal{F}_i$ and the dense text encoder $\mathcal{F}^{d}_t$ map images and dense textual descriptions into a shared embedding space.  Formally, given a dense textual description $T^d$ and a raw image as $I$, our goal is to map these inputs into compact vectors $\mathbf{t}^d \in \mathbb{R}^d$ and $\mathbf{i} \in \mathbb{R}^d$ such that their semantic relationships can be quantified through a similarity metric $S(\cdot, \cdot)$. Specifically, the similarity function is formulated as:

\begin{equation}
    \begin{aligned}
        S(I, T) = \frac{\mathbf{i}^T \mathbf{t}^d}{||\mathbf{i}||\cdot ||\mathbf{t}^d||}
    \end{aligned}
\end{equation}

To optimize the model, we use triplet loss function with the hardest negative samples to minimize the similarity between relevant pairs and to maximize the distance between irrelevant ones. 
The optimized loss function is mathematically formulated as:
\begin{equation}
    \begin{aligned}
        \mathcal{L}_{pretrain} =  &\sum_{(I,T^d)\in \mathcal{D}}\left[\alpha - S(I,T^d) + S(I,\hat{T^d})\right]_+\\
    &+ \left[\alpha - S(I,T^d) + S(\hat{I},T^d)\right]_+
    \end{aligned}
\end{equation}
where $\alpha$ serves as a margin parameter and $[x]_+ \equiv Max(x,0)$.
In dataset $\mathcal{D}$, the similarity in a positive pair $S\left(I, T^d\right)$ should be higher than that in the hardest negative pairs $S( \hat{I}, {T^d})$ and $S({I}, \hat{{T^d}})$ by a margin $\alpha$.

\begin{figure}
    \centering
    \includegraphics[width=\linewidth]{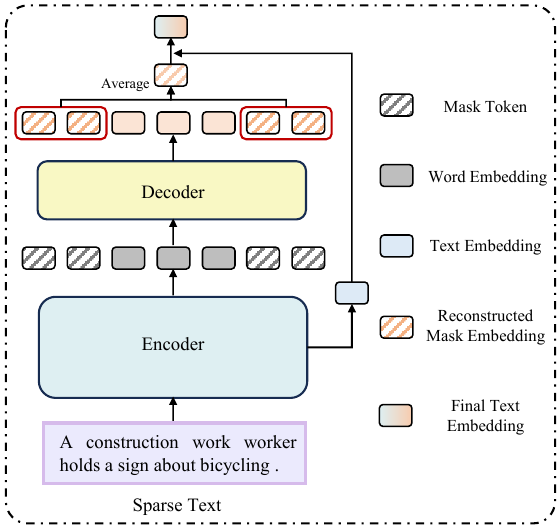}
    \caption{Architecture for the sparse text branch. Similar to the decoder in MAE \cite{he2022masked}, we introduce a set of learnable tokens in the sparse text branch to predict the implicit context of sparse text through the decoder. The learnable tokens help in reconstructing the missing  semantics, allowing the model to better capture the underlying context and relationships within the sparse text.}
    \label{fig:text_branch}
\end{figure}

\begin{table*}[!t]
\small

\centering
\setlength{\tabcolsep}{0.4mm}
\begin{tabular}{lcccccccccccccccccccccc}
\topline

\headcol&&
\multicolumn{7}{c}{{\textbf{Flickr30k 1K Test}}}&
\multicolumn{7}{c}{{\textbf{MS-COCO 5-fold 1K Test}}} & 
\multicolumn{7}{c}{{\textbf{MS-COCO 5K Test}}} \\ 
\arrayrulecolor{tableheadcolor}
\specialrule{6pt}{0pt}{-6pt}
\arrayrulecolor{black}
\cmidrule[0.5pt](lr){3-9} 
\cmidrule[0.5pt](lr){10-16} 
\cmidrule[0.5pt](lr){17-23} 
\headcol\hspace{3em}& &
\multicolumn{3}{c}{\textbf{Text   Retrieval}}  & 
\multicolumn{3}{c}{\textbf{Image   Retrieval}}  & 
 &
\multicolumn{3}{c}{\textbf{Text   Retrieval}}  & 
\multicolumn{3}{c}{\textbf{Image   Retrieval}}&
&
\multicolumn{3}{c}{\textbf{Text   Retrieval}}  & 
\multicolumn{3}{c}{\textbf{Image   Retrieval}}& 
\\ 
\arrayrulecolor{tableheadcolor}
\specialrule{6pt}{0pt}{-6pt}
\arrayrulecolor{black}\cmidrule[0.5pt](lr){3-5} \cmidrule[0.5pt](lr){6-8} \cmidrule[0.5pt](lr){10-12} \cmidrule[0.5pt](lr){13-15} \cmidrule[0.5pt](lr){17-19} \cmidrule[0.5pt](lr){20-22}

\headcol \multirow{-2}{*}{\cellcolor{tableheadcolor}\textbf{Method}}&
\multirow{-2}{*}{\cellcolor{tableheadcolor}\textbf{FG}}&
R1& R5& R10 & R1& R5& R10& \multirow{-2}{*}{rSum}&
R1& R5& R10 & R1& R5& R10& \multirow{-2}{*}{rSum}&
R1& R5& R10 & R1& R5& R10& \multirow{-2}{*}{rSum}\\ 
\arrayrulecolor{tableheadcolor}
\specialrule{6pt}{0pt}{-2pt}
\arrayrulecolor{black}
\midrule[0.75pt]

\multicolumn{20}{l}{\textbf{\textit{ViT-Base-224 + BERT-base}}}\\ 
VSE++ \cite{faghri2017vse++} &\XSolidBrush&
71.8&92.8&96.5&59.4&84.7&90.9&496.1&
75.0&94.6&98.0&62.7&89.4&94.9&514.6&
62.3&87.6&93.4&43.9&73.6&83.3&413.4\\		
SCAN \cite{lee2018stacked}&\Checkmark&
69.5&90.9&95.6&56.4&83.1&90.0&485.6&
76.0&95.4&98.1&64.5&90.8&95.8&520.6&
53.9&81.8&90.0&42.9&72.3&82.5&423.5
\\
SGR \cite{diao2021similarity}&\Checkmark&
69.7&90.8&95.2&59.1&84.1&89.9&488.7&
77.2&95.0&98.0&65.1&90.7&95.8&521.8&
54.9& 82.8& 90.5& 42.8& 72.2& 82.5&425.8
\\
CHAN \cite{pan2023fine}&\Checkmark&
69.2&91.8&95.0&58.4&84.9&90.6&489.9&
77.1&95.1&98.1&65.0&91.0&96.0&522.2&
56.3& 83.2& 90.1& 43.0& 72.6& 82.8&428.0
\\
LAPS \cite{fu2024linguistic} &\Checkmark&
74.0&93.4&97.4&62.5&87.3&92.7&507.3&
78.7&95.5&98.3&66.2&91.3&96.2&526.3&
57.5& 84.0& 90.8& 44.5& 74.0& 83.6&434.4
\\
AVSE \cite{liu2025asymmetric} &\XSolidBrush&
{76.0}&{94.6}&{97.5}&{62.7}&{88.4}&{93.1}&512.3 &
{79.8}&{95.6}&{98.3}&{67.0}& {91.5}& {96.3}& 528.5&
{58.8}&{84.3}&{91.0}&{45.1}&{74.3}&{83.9}&437.4
\\
\rowcol \textbf{D2S-VSE} &\XSolidBrush&
\textbf{82.8}&\textbf{96.1}&\textbf{98.3}&\textbf{68.5}&\textbf{91.3}&\textbf{94.9}&\textbf{531.9}&
\textbf{80.1}&\textbf{97.0}&\textbf{99.2}&\textbf{68.1}&\textbf{92.5}&\textbf{96.7}&\textbf{533.7}&
\textbf{60.1}&\textbf{85.5}&\textbf{92.5}&\textbf{46.3}&\textbf{75.9}&\textbf{85.2}&\textbf{445.6}
\\
\midrule[0.5pt]
\multicolumn{15}{l}{\textbf{\textit{ViT-Base-384 + BERT-base}}}\\ 
VSE++ \cite{faghri2017vse++} &\XSolidBrush&
77.1&95.7&97.5&65.8&90.2&94.3&520.5&
77.0&95.7&98.4&64.6&91.1&96.2&523.0&
54.9& 82.8& 90.4& 42.4& 72.4& 82.8&425.8
\\	
SCAN \cite{lee2018stacked}&\Checkmark&
75.4&94.4&96.9&63.6&88.6&93.5&512.5&
76.1&95.5&98.5&65.1&91.6&96.3&523.1&
53.3&81.8& 90.0& 42.6& 72.6& 82.9&423.1
\\
SGR \cite{diao2021similarity}&\Checkmark&
76.9&94.9&98.1&64.2&88.4&93.3&515.8&
75.8&95.7&98.6&65.6&92.0&96.5&524.2&
53.3& 81.0& 89.6& 42.9& 73.1& 83.7&423.6
\\
CHAN \cite{pan2023fine}&\Checkmark&
75.4&94.5&97.6&63.2&88.6&93.1&512.4&
78.1&95.8&98.6&66.1&82.1&96.6&527.3&
55.6& 83.8& 91.2& 43.4& 73.6& 83.5&431.1
\\
LAPS \cite{fu2024linguistic} &\Checkmark&
79.0&96.0&98.1&67.3&90.5&94.5&525.4&
78.6&96.3&98.9&68.0&92.4&96.8&531.0&
57.4& 84.9& 92.5&46.4&75.8&85.2&442.2
\\
AVSE \cite{liu2025asymmetric}&\XSolidBrush&
{80.3}&{96.4}&{98.7}&{67.9}&{91.2}&{94.7}&529.2&
\textbf{81.1}&{97.1}&{99.0}&{68.3}& {92.7}& \textbf{97.0}&535.2&
\textbf{61.2}&\textbf{86.8}&\textbf{93.2}&46.2&{75.9}&85.0&448.3
\\
\rowcol \textbf{D2S-VSE} &\XSolidBrush&
\textbf{84.1}&\textbf{97.5}&\textbf{99.1}&\textbf{70.3}&\textbf{91.6}&\textbf{95.3}&\textbf{537.9}&
80.8&\textbf{97.2}&\textbf{99.1}&\textbf{69.0}&\textbf{92.9}&{96.8}&\textbf{535.8}&
60.6&86.5&\textbf{93.2}&\textbf{46.8}&\textbf{76.4}&\textbf{85.7}&\textbf{449.1}
\\
\bottomrule[1pt]
\end{tabular}
\caption{
   Comparison of experimental results using ViT-Base \cite{dosovitskiy2020image} and BERT \cite{devlin2018bert} the MS-COCO and Flickr30K datasets. ``FG” indicates whether fine-grained cross-modal alignment is applied.
    The best performance is highlighted in \textbf{bold}.
}
\label{table:t1}
\end{table*}

\subsection{Finetune With Sparse Description}
\label{sec:3.2}
Stage \uppercase\expandafter{\romannumeral1} pre-training with dense descriptions enhances the information capacity of embeddings of the VSE model,
but there exists a gap between the training and testing data, as the model is trained on dense texts and tested on sparse texts, leading to suboptimal performance.
However, simply fine-tuning directly on the image-sparse text pair data will not make the sparse text embedding contain richer semantics, that is, more information capacity.
To address this problem, we introduce a novel distillation mechanism while aligning, which treats sparse text as masked dense text and designs a decoder. Through distillation, the decoder can decode sparse text embeddings into dense text embeddings, thereby improving the information capacity of sparse text embeddings.
This mechanism extracts sparse text embeddings from dense text embeddings while aligning images and text, thereby enhancing the information capacity of sparse text features.

\subsubsection{Aligning Information Capacity by Dense-to-Sparse Feature Distillation}
To ensure that the sparse text embedding retains as much information as the dense text embedding, we introduce a feature distillation mechanism from dense to sparse.
Inspired by masked signal modeling \cite{he2022masked, devlin2018bert}, we treat sparse text as a masked version of dense text. This allows us to decode and reconstruct the sparse text embedding through feature distillation, enabling the reconstructed embedding to approximate the information capacity of the dense text embedding.

\noindent \textbf{Transformer-based Decoder.} To achieve this reconstruction, we design a Transformer-based decoder, as illustrated in Figure \ref{fig:text_branch}. 
The decoder plays a pivotal role in D2S-VSE, responsible for reconstructing the dense text embedding from the sparse text embedding by leveraging feature distillation. It learns to restore the missing contextual information in sparse text, ensuring that the final sparse text embedding retains as much semantic richness as the dense text embedding.
Given a sparse text $T^s$, we first feed it into the sparse text encoder $\mathcal{F}^s_t$ to obtain the word embedding $\mathbf{w}^s$ and the sparse text embedding $\mathbf{t}^s$.
Next, learnable mask tokens with position $\mathbf{m}$ are concatenated with $\mathbf{w}^s$ and input into the decoder for processing. The decoder outputs the reconstructed tokens $\bar{\mathbf{m}}$, which are then averaged to obtain $\bar{\mathbf{t}}$.
Finally, we sum $\mathbf{t}^s$ and $\bar{\mathbf{t}}$ to generate the final sparse text embedding $\hat{\mathbf{t}^s}$. This process can be formulated as:
\begin{equation}
    \hat{\mathbf{t}^s} = Decoder([\mathbf{w}^s, \mathbf{m}]) + \mathbf{t}^s
\end{equation}

This reconstruction process enables the mask tokens to capture the contextual information missing from the sparse text, aligning it more closely with the dense text embedding. By combining the two, we obtain a more comprehensive and enriched text feature embedding.

\noindent \textbf{Dense-to-Sparse Distillation.}
Through the dense text pre-training in stage 1, the representation of the pre-trained VSE model can capture richer semantics, and the pre-trained dense text encoder can be used as a teacher to distill the information capacity of dense text embeddings into sparse text embeddings.
This Dense-to-Sparse distillation minimizes a negative cosine similarity between then dense text embedding $\mathbf{t}^d$ and the final sparse text embedding $\hat{\mathbf{t}^s}$:
\begin{equation}
    \begin{aligned}
        \mathcal{L}_{distill}(\mathbf{t}^d, \hat{\mathbf{t}^s}) = 1 - \frac{\mathbf{t}^d \cdot \hat{\mathbf{t}^s}}{||\mathbf{t}^d|| \cdot ||\hat{\mathbf{t}^s}||}
    \end{aligned}
\end{equation}

\subsubsection{Aligning Visual Semantic Using Sparse text}
At the same time, we also fine-tune the VSE model on image-sparse text pair data to adapt the model to common sparse text queries in cross-modal retrieval.
Given image $I$ and sparse text $T^s$, we extract image embedding $\mathbf{i}$ and sparse text embedding $\mathbf{t}$ and optimize the VSE model as follow:
\begin{equation}
    \begin{aligned}
        \mathcal{L}_{align} =  &\sum_{(I,T^s)\in \mathcal{D}}\left[\alpha - S(I,T^s) + S(I,\hat{T^s})\right]_+\\
    &+ \left[\alpha - S(I,T^s) + S(\hat{I},T^s)\right]_+
    \end{aligned}
\end{equation}

In summary, the final loss function is defined as follows to perform joint optimization of the two objectives.
\begin{equation}
    \mathcal{L} = \mathcal{L}_{align}  + \mathcal{L}_{distill} 
\label{loss}
\end{equation}

\section{Experiments}

\subsection{Dataset and Settings}
\noindent {\bf Datasets.}
Following previous works \cite{faghri2017vse++}, we conduct our experiments on two widely used benchmark datasets: MS-COCO \cite{lin2014microsoft} and Flickr30K \cite{plummer2015flickr30k}.MS-COCO consists of 123,287 images, with each image annotated with five text captions. Following the dataset split in \cite{faghri2017vse++}, we use 113,287 images for training, 5,000 for validation, and 5,000 for testing.
Flickr30K contains 31,783 images, each paired with five textual descriptions. We adopt the standard split from \cite{faghri2017vse++}, using 29,000 images for training, 1,000 for validation, and 1,000 for testing.

\noindent {\bf Evaluation Metrics.}
For image-text matching, we use the commonly adopted Recall@K (where K = 1, 5, 10), denoted as R@1, R@5, and R@10. These metrics represent the percentage of ground-truth instances retrieved within the top 1, 5, or 10 results, respectively.

\begin{table}[]

\center
\setlength{\tabcolsep}{0.9mm}
\small
\begin{tabular}{lccccccc}
\topline
\headcol  &\multicolumn{3}{c}{\bf{Text   Retrieval}} &\multicolumn{3}{c}{\bf{Image   Retrieval}}&  \\ 
\arrayrulecolor{tableheadcolor}
\specialrule{6pt}{0pt}{-6pt}
\arrayrulecolor{black}
\cmidrule(lr){2-4} \cmidrule(lr){5-7}
\headcol \multirow{-2}{*}{\bf{Method}}& R@1  & R@5  & R@10  & R@1  & R@5  & R@10 &\multirow{-2}{*}{{rSum}} \\ 
\arrayrulecolor{tableheadcolor}
\specialrule{6pt}{0pt}{-3pt}
\arrayrulecolor{black}
\midrule[0.75pt]
\multicolumn{8}{l}{\textbf{Pre-trained ResNet-152 on IN + GRU}}
  \\ 

  VSE$\infty$ 
\cite{chen2021learning} & 
 76.5& 94.2& {97.7}& 56.4& 83.4& 89.9&498.1
 \\
MV-VSE
\cite{li2022multi} & 
79.0 &94.9 &97.7 &59.1 &84.6 &90.6 &505.9
\\
NAAF* 
\cite{zhang2022negative} & 
{81.9}  &96.1 &98.3 &61.0 &85.3 &90.6 &513.2
\\
HREM* \cite{fu2023learning} 
& {81.4}& {96.5}& {98.5}& {60.9}& {85.6}& {91.3}&514.2
\\
CHAN \cite{pan2023fine}
& {79.7}& {94.5}& {97.3}& {60.2}& {85.3}& {90.7}& 507.7
\\
AVSE \cite{liu2025asymmetric} &
{82.3} &{96.2} &{98.5} &\textbf{{62.8}} &{88.0} &{92.8} &520.6 \\
\rowcol \textbf{D2S-VSE} &\textbf{82.5}&\textbf{96.6}&\textbf{99.0}&62.6&\textbf{88.1}&\textbf{93.1}&\textbf{521.9}
\\
\midrule[0.5pt]
\multicolumn{7}{l}{\textbf{\textit{Swin-Base-224 + BERT-base}}}\\ 
VSE++ \cite{faghri2017vse++} &
82.5&96.5&98.9&70.0&91.4&95.1&534.4

\\	
SCAN \cite{lee2018stacked}&
79.0&95.9&98.2&67.7&90.6&94.9&526.3
\\
SGR \cite{diao2021similarity}&
80.4&97.0&98.7&66.9&90.2&94.5&527.6
\\
CHAN \cite{pan2023fine})&
81.4&97.0&98.6&68.5&90.6&94.5&530.6
\\
LAPS\cite{fu2024linguistic} &
82.4&97.4&{99.5}&70.0&91.7&95.4&536.3
\\
AVSE \cite{liu2025asymmetric}  &
{83.9}&{97.4}&99.4&{70.0}&{92.4}&{95.6}&538.7
\\
\rowcol \textbf{D2S-VSE} &
\textbf{87.2}&\textbf{98.4}&\textbf{99.9}&\textbf{73.0}&\textbf{93.5}&\textbf{96.7}&\textbf{548.7}
\\
\midrule[0.5pt]
\multicolumn{7}{l}{\textbf{\textit{Swin-Base-384 + BERT-base}}}\\ 
VSE++ \cite{faghri2017vse++} &
83.3&97.5&99.2&71.1&93.2&96.2&540.6
\\	
SCAN \cite{lee2018stacked}&
81.9&96.9&98.9&70.0&92.7&95.8&536.1
\\
SGR \cite{diao2021similarity}&
80.7&96.8&99.0&69.9&91.7&95.3&533.4
\\
CHAN \cite{pan2023fine}&
81.2&96.7&98.8&70.3&92.2&95.9&535.0

\\
LAPS \cite{fu2024linguistic} &
85.1&97.7&99.2&{74.0}&93.0&96.3&545.3
\\
AVSE \cite{liu2025asymmetric} &
{87.1}&{98.3}&{99.2}&73.6&{93.5}&{96.5}&548.2

\\
\rowcol \textbf{D2S-VSE} &
\textbf{87.8}&\textbf{99.0}&\textbf{99.7}&\textbf{75.7}&\textbf{94.1}&\textbf{96.9}&\textbf{553.2}

\\
\bottomrule[1pt]
\end{tabular}
\caption{
    Comparison of experimental results using other common backbones on the Flickr30K dataset. The best performance is highlighted in \textbf{bold}.
    `*' indicates ensemble methods.
}
\label{table:t2}
\end{table}

\subsection{Implementation details}
\label{details}
For dense text generation, we use LLaVa \cite{liu2023visual} to generate dense textual descriptions. The hyperparameters for LLaVa are set as follows: Top-P $= 0.9$, temperature$ = 0.2$, and max-new-tokens$ = 500$. We generate a dense text description for each image in the MS-COCO and Flickr30K datasets, with each dense text sentence corresponding to $\frac{1}{5}$ of the sparse text descriptions.

In the pre-training stage, we use the AdamW optimizer \cite{loshchilov2019decoupled} to train the VSE model for 30 epochs, with a mini-batch size of 128. 
The learning rate is initially set to 0.0005 for the first few epochs and is reduced by a factor of 10 during the 9th, 15th, 20th, and 25th epochs. 
In the fine-tuning stage, we freeze the dense text encoder and fine-tune both the image encoder and sparse text encoder. We train them using the same training strategy as in the pre-training stage and evaluate the best model on the validation set.

For model details, we use traditional region features and employ the recently popular Vision Transformer (ViT) \cite{dosovitskiy2020image} and Swin Transformer \cite{liu2021swin} as the backbone. We set the dimension of the shared embedding space d to 512.
For feature aggregation, we use GPO \cite{chen2021learning} for both the image encoder and the text encoder. The decoder consists of 4 layers of Transformer, with 4 attention heads per layer. We set 100 learnable tokens to predict the context of sparse text embeddings. \textit{We provide a more detailed implementation setup in the supplementary material.}

\begin{figure*}
  \centering
  \begin{subfigure}{0.33\linewidth}
    \centering
    \includegraphics[width=\linewidth]{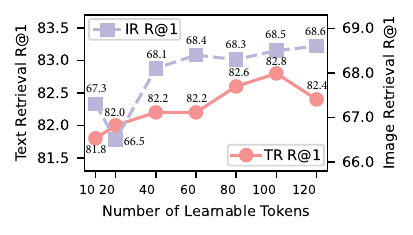}
    \caption{The impact of Learnable Tokens}
    \label{fig:subfig1}
  \end{subfigure}
  \begin{subfigure}{0.33\linewidth}
    \centering
    \includegraphics[width=\linewidth]{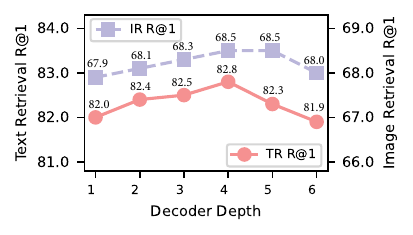}
    \caption{The impact of Decoder Depth}
    \label{fig:subfig2}
  \end{subfigure}
  \begin{subfigure}{0.33\linewidth}
    \centering
    \includegraphics[width=\linewidth]{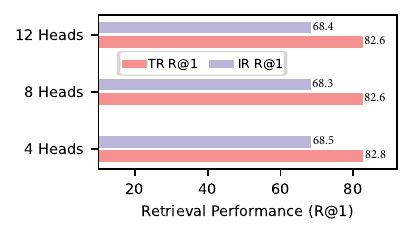}
    \caption{The impact of Decoder Head}
    \label{fig:subfig3}
  \end{subfigure}
  \caption{Ablation study of the architecture of decoder on Flickr30K. If not specified, the default is: the number of the learnable tokens is 100, the decoder has 4 layers and 4 head.}
  \label{fig:decoder}
\end{figure*}

\subsection{Comparisons with Recent State-of-the-arts}
We compare our D2S-VSE with the most recent state-of-the-art approaches on two widely used datasets, i.e., Flirkr30K and MS-COCO. 
For the sake of fairness, our D2S-VSE employs a diverse range of backbones to facilitate comparisons with various previous methods.

In Table \ref{table:t1}, we present the results of using ViT-base \cite{dosovitskiy2020image} + BERT-base \cite{devlin2018bert} as the backbone on the Flickr30K and MS-COCO datasets. Our D2S-VSE surpasses all state-of-the-art methods, demonstrating remarkable performance.
In this setting, we compare our method against the two leading approaches, LAPS \cite{fu2024linguistic} and AVSE \cite{liu2025asymmetric}, and achieve superior results across nearly all metrics. For instance, on the Flickr30K dataset, our approach outperforms AVSE in text retrieval, achieving a 6.8\% improvement in R@1 with the ViT-base-224 backbone and a 3.8\% improvement with the ViT-base-384 backbone.
Similarly, for image retrieval, our method achieves gains of 5.8\% and 2.4\% when using the ViT-base-224 and ViT-base-384 backbones, respectively. A consistent performance boost is also observed across all metrics on the MS-COCO dataset.

Additionally, we conducted experiments using other commonly used backbones, as shown in Table \ref{table:t2}. Specifically, we evaluated our approach on the Flickr30K dataset with Faster R-CNN \cite{anderson2018bottom} + GRU and Swin-Base \cite{liu2021swin} + BERT-Base as backbones. The results further confirm that our method consistently outperforms existing approaches across different backbone architectures.

\subsection{Ablation Study}
To verify the effectiveness of each component of our D2S-VSE, we conduct extensive ablation studies on the Flickr30K dataset. All experiments are performed using ViT-base-224 and BERT-base as the backbone to ensure a consistent and fair comparison.

\begin{table}[]

\center
\setlength{\tabcolsep}{1mm}
\scriptsize
\begin{tabular}{cccccccccc}
\topline
\headcol &\multicolumn{2}{c}{\bf{Finetune}}&\multicolumn{3}{c}{\bf{Text   Retrieval}} &\multicolumn{3}{c}{\bf{Image   Retrieval}}&  \\ 
\arrayrulecolor{tableheadcolor}
\specialrule{6pt}{0pt}{-6pt}
\arrayrulecolor{black}
\cmidrule(lr){2-3}\cmidrule(lr){4-6} \cmidrule(lr){7-9}
\headcol \multirow{-2}{*}{\bf{Pretrain}}& \textbf{Align} & \textbf{Distill}&R@1  & R@5  & R@10  & R@1  & R@5  & R@10 &\multirow{-2}{*}{{rSum}} \\ 
\arrayrulecolor{tableheadcolor}
\specialrule{6pt}{0pt}{-3pt}
\arrayrulecolor{black}
\midrule[0.75pt]
 \Checkmark & & &56.2&84.7&91.6&41.7&71.3&79.7&425.2\\
 & \Checkmark & &76.3&93.9&97.3&61.2&87.6&93.2&509.5\\
\Checkmark &\Checkmark &&79.2&96.5&98.3&66.8&90.5&94.4&525.7\\
\Checkmark & &\Checkmark&76.0&94.5&97.3&62.9&88.3&93.1&512.1\\
\rowcol \Checkmark &\Checkmark &\Checkmark&\textbf{82.8}&\textbf{96.1}&\textbf{98.3}&\textbf{68.5}&\textbf{91.3}&\textbf{94.9}&\textbf{531.9}
\\
\bottomrule[1pt]
\end{tabular}
\caption{
   Ablation experiment on the impact of our framework on Flick30K using ViT-base-224 + BERT-base as backbones. The default setting is highlighted in \textcolor{Lavender!50}{Light Blue}.
}
\label{table:t3}
\end{table}

\noindent \textbf{The impact of the two-stage training framework.}
Our D2S-VSE introduces a two-stage training framework to enhance the information capacity of image embedding and text embedding, and achieves very good results on commonly used datasets. In order to explore the role of each component in the framework, we conducted ablation experiments as shown in Table \ref{table:t3}.

\ul{\textit{Is the model pre-trained with dense text inherently superior, leading to better performance?}} To verify this, we directly tested the pre-trained model on the Flickr30K test set. We found that due to the gap between dense and sparse text, the pre-trained model struggles to adapt to sparse text, resulting in rSum of only 425.2 on the Flickr30K test set.

\ul{\textit{The pre-training stage enhances the information capacity of image embeddings!}}
We observed that the R@1 of text retrieval improved by 2.9\% after pre-training compared to the model without pre-training. This suggests that dense text pre-training enhances the information capacity of image embeddings, enabling the model to better align multi-view descriptions.

\begin{figure*}
    \centering
    \includegraphics[width=0.8\linewidth]{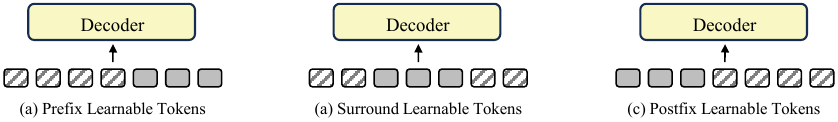}
    \caption{Different type of the position strategy of learnable tokens. Tokens at different positions can predict the semantics of different positions in dense text.}
    \label{fix}
\end{figure*}

\ul{\textit{Dense-to-Sparse feature distillation enhances the information capacity of sparse text embeddings!}}
Applying dense-to-sparse distillation while aligning images during the fine-tuning stage further boosts the model's retrieval performance. On the Flickr30K dataset, text retrieval and image retrieval reached 82.8\% and 68.5\%, respectively. This improvement stems from distillation enhances the information capacity of text embeddings, leading to better semantic consistency between image and text embeddings.

\textit{Different type of distillation functions.} We used L1 distance, L2 distance, and negative cosine similarity as the loss functions for Dense-to-Sparse distillation, respectively. As shown in Table \ref{table:t4}, the effect of negative cosine similarity distillation is much better than the other two.

\noindent\textbf{Decoder design.} The decoder is the most important part of Dense-to-Sparse distillation. Its existence allows learnable tokens to predict possible contexts based on the word embeddings of sparse text, thereby increasing the information capacity of sparse text features. As shown in Figure \ref{fig:decoder}, we conduct ablation studies of Learnable tokens, decoder depth, and decoder head.

\ul{\textit{The number of tokens determines the reconstruction capability of the decoder.}} As shown in Figure \ref{fig:subfig1}, we observed that as the number of learnable tokens in the decoder increases, the model's performance in both image retrieval and text retrieval gradually improves, stabilizing when the number of tokens reaches approximately 100. At this point, the R@1 for text retrieval and image retrieval reaches 82.8\% and 68.5\%, respectively. However, when the number of tokens increases to 120, the model shows no significant advantage. Therefore, considering both performance and efficiency, we set the number of tokens to 100.

\begin{table}[]

\center
\setlength{\tabcolsep}{0.9mm}
\scriptsize
\begin{tabular}{lccccccc}
\topline
\headcol &\multicolumn{3}{c}{\bf{Text   Retrieval}} &\multicolumn{3}{c}{\bf{Image   Retrieval}}&  \\ 
\arrayrulecolor{tableheadcolor}
\specialrule{6pt}{0pt}{-6pt}
\arrayrulecolor{black}
\cmidrule(lr){2-4} \cmidrule(lr){5-7}
\headcol \multirow{-2}{*}{\bf{Method}}& R@1  & R@5  & R@10  & R@1  & R@5  & R@10 &\multirow{-2}{*}{{rSum}} \\ 
\arrayrulecolor{tableheadcolor}
\specialrule{6pt}{0pt}{-3pt}
\arrayrulecolor{black}
\midrule[0.75pt]
$\ell_1$ Distance &81.9&95.3&98.2&68.0&90.8&94.7&528.9\\
$\ell_2$ Distance &82.1&95.6&\textbf{98.4}&67.9&91.2&\textbf{95.1}&530.2\\
\rowcol Negative Cosine Similarity &\textbf{82.8}&\textbf{96.1}&98.3&\textbf{68.5}&\textbf{91.3}&94.9&\textbf{531.9}
\\
\bottomrule[1pt]
\end{tabular}
\caption{
   Ablation experiment on the impact of different distillation functions on Flick30K using ViT-base-224 + BERT-base as backbones. The default setting is highlighted in \textcolor{Lavender!50}{Light Blue}. 
}
\label{table:t4}
\end{table}

\ul{\textit{Surrounding learnable tokens can better predict the context of sparse text.}} In addition, we conducted an ablation experiment on the insertion position of learnable tokens. Since our motivation stems from treating sparse text as masked dense text, with learnable tokens used to reconstruct the masked semantics, their placement is crucial. As shown in Figure \ref{fix}, we designed Prefix Learnable Tokens, Surround Learnable Tokens, and Postfix Learnable Tokens to explore the impact of token positioning on model performance.
Table \ref{table:t5} presents the ablation results on the impact of different learnable token positions. We observe that the rSum index of Surround Learnable Tokens is 2.6\% and 2.4\% higher than that of Prefix Learnable Tokens and Postfix Learnable Tokens, respectively. This outcome is intuitive, as the exact location of sparse text semantics within dense text is unknown. The Surround insertion method allows the model to predict semantics from both the preceding and following contexts simultaneously, leading to improved performance.

\begin{table}[]
\center
\scriptsize
\begin{tabular}{lccccccc}
\topline
\headcol &\multicolumn{3}{c}{\bf{Text   Retrieval}} &\multicolumn{3}{c}{\bf{Image   Retrieval}}&  \\ 
\arrayrulecolor{tableheadcolor}
\specialrule{6pt}{0pt}{-6pt}
\arrayrulecolor{black}
\cmidrule(lr){2-4} \cmidrule(lr){5-7}
\headcol \multirow{-2}{*}{\bf{Type}}& R@1  & R@5  & R@10  & R@1  & R@5  & R@10 &\multirow{-2}{*}{{rSum}} \\ 
\arrayrulecolor{tableheadcolor}
\specialrule{6pt}{0pt}{-3pt}
\arrayrulecolor{black}
\midrule[0.75pt]
Prefix &81.7&95.3&98.2&68.1&91.2&94.8&529.3\\
Postfix &82.0&95.6&\textbf{98.3}&67.8&91.1&{94.7}&529.5\\
\rowcol Surround &\textbf{82.8}&\textbf{96.1}&\textbf{98.3}&\textbf{68.5}&\textbf{91.3}&\textbf{94.9}&\textbf{531.9}
\\
\bottomrule[1pt]
\end{tabular}
\caption{
   Ablation experiment on the impact of the position of the learnable tokens on Flick30K using ViT-base-224 + BERT-base as backbones. The default setting is highlighted in \textcolor{Lavender!50}{Light Blue}. 
}
\label{table:t5}
\end{table}

\ul{\textit{ A deep decoder can improve retrieval performance.}} As shown in Figure \ref{fig:subfig2}, as the decoder depth increases, retrieval performance first improves and then declines, with the best results achieved when the number of layers is set to 4.
Although our approach is inspired by Masked Signal Modeling \cite{he2022masked}, our findings regarding the decoder differ. In Masked Signal Modeling, the objective is to enhance the encoder through reconstruction, where an excessively deep decoder may hinder the encoder’s capability. However, in our D2S-VSE, retrieval performance relies on the predictive ability of the decoder, meaning that a deeper decoder generally leads to better results.

\ul{\textit{Our D2S-VSE can work with very few decoder heads.}} As shown in Figure \ref{fig:subfig3}, the number of attention heads has minimal impact on performance, with only 4 heads required to achieve optimal results. Increasing the number of heads does not lead to significant improvements, as all performance differences remain within 0.2\%.

\section{Conclutions}
In this paper, we highlight the importance of embedding information capacity in addressing the multi-view description problem in image-text matching tasks. To tackle this challenge, we propose a novel Visual Semantic Embedding (VSE) framework, Dense-to-Sparse Feature Distillation for Visual Semantic Embedding (D2S-VSE). The core idea of D2S-VSE is to enhance the information capacity of visual semantic embeddings by aligning dense text with images during pre-training. In the fine-tuning stage, we further refine sparse text embeddings through Dense-to-Sparse feature distillation, effectively improving their information capacity and leading to better image-text matching performance. Comprehensive experiments on two widely-used benchmarks validate the effectiveness of the proposed method, leading to state-of-the-art performance.
{
    \small
    \bibliographystyle{ieeenat_fullname}
    \bibliography{main}

\begin{thebibliography}{41}
\providecommand{\natexlab}[1]{#1}
\providecommand{\url}[1]{\texttt{#1}}
\expandafter\ifx\csname urlstyle\endcsname\relax
  \providecommand{\doi}[1]{doi: #1}\else
  \providecommand{\doi}{doi: \begingroup \urlstyle{rm}\Url}\fi

\bibitem[Anderson et~al.(2018)Anderson, He, Buehler, Teney, Johnson, Gould, and Zhang]{anderson2018bottom}
Peter Anderson, Xiaodong He, Chris Buehler, Damien Teney, Mark Johnson, Stephen Gould, and Lei Zhang.
\newblock Bottom-up and top-down attention for image captioning and visual question answering.
\newblock In \emph{Proc. IEEE/CVF Conf. Comput. Vis. Pattern Recognit. (CVPR)}, pages 6077--6086, 2018.

\bibitem[Chen et~al.(2020)Chen, Ding, Liu, Lin, Liu, and Han]{chen2020imram}
Hui Chen, Guiguang Ding, Xudong Liu, Zijia Lin, Ji Liu, and Jungong Han.
\newblock Imram: Iterative matching with recurrent attention memory for cross-modal image-text retrieval.
\newblock In \emph{Proc. IEEE/CVF Conf. Comput. Vis. Pattern Recognit. (CVPR)}, pages 12655--12663, 2020.

\bibitem[Chen et~al.(2021)Chen, Hu, Wu, Jiang, and Wang]{chen2021learning}
Jiacheng Chen, Hexiang Hu, Hao Wu, Yuning Jiang, and Changhu Wang.
\newblock Learning the best pooling strategy for visual semantic embedding.
\newblock In \emph{Proc. IEEE/CVF Conf. Comput. Vis. Pattern Recognit. (CVPR)}, pages 15789--15798, 2021.

\bibitem[Chun(2023)]{chun2023improved}
Sanghyuk Chun.
\newblock Improved probabilistic image-text representations.
\newblock \emph{arXiv preprint arXiv:2305.18171}, 2023.

\bibitem[Chun et~al.(2021)Chun, Oh, De~Rezende, Kalantidis, and Larlus]{chun2021probabilistic}
Sanghyuk Chun, Seong~Joon Oh, Rafael~Sampaio De~Rezende, Yannis Kalantidis, and Diane Larlus.
\newblock Probabilistic embeddings for cross-modal retrieval.
\newblock In \emph{CVPR}, pages 8415--8424, 2021.

\bibitem[Devlin et~al.(2018)Devlin, Chang, Lee, and Toutanova]{devlin2018bert}
Jacob Devlin, Ming-Wei Chang, Kenton Lee, and Kristina Toutanova.
\newblock Bert: Pre-training of deep bidirectional transformers for language understanding.
\newblock \emph{arXiv preprint arXiv:1810.04805}, 2018.

\bibitem[Diao et~al.(2021)Diao, Zhang, Ma, and Lu]{diao2021similarity}
Haiwen Diao, Ying Zhang, Lin Ma, and Huchuan Lu.
\newblock Similarity reasoning and filtration for image-text matching.
\newblock In \emph{Proc. AAAI Conf. Artif. Intell.}, pages 1218--1226, 2021.

\bibitem[Dosovitskiy et~al.(2020)Dosovitskiy, Beyer, Kolesnikov, Weissenborn, Zhai, Unterthiner, Dehghani, Minderer, Heigold, Gelly, et~al.]{dosovitskiy2020image}
Alexey Dosovitskiy, Lucas Beyer, Alexander Kolesnikov, Dirk Weissenborn, Xiaohua Zhai, Thomas Unterthiner, Mostafa Dehghani, Matthias Minderer, Georg Heigold, Sylvain Gelly, et~al.
\newblock An image is worth 16x16 words: Transformers for image recognition at scale.
\newblock \emph{arXiv preprint arXiv:2010.11929}, 2020.

\bibitem[Faghri et~al.(2018)Faghri, Fleet, Kiros, and Fidler]{faghri2017vse++}
Fartash Faghri, David~J Fleet, Jamie~Ryan Kiros, and Sanja Fidler.
\newblock Vse++: Improving visual-semantic embeddings with hard negatives.
\newblock In \emph{BMVC}, 2018.

\bibitem[Fu et~al.(2023)Fu, Mao, Song, and Zhang]{fu2023learning}
Zheren Fu, Zhendong Mao, Yan Song, and Yongdong Zhang.
\newblock Learning semantic relationship among instances for image-text matching.
\newblock In \emph{Proc. IEEE/CVF Conf. Comput. Vis. Pattern Recognit. (CVPR)}, pages 15159--15168, 2023.

\bibitem[Fu et~al.(2024)Fu, Zhang, Xia, and Mao]{fu2024linguistic}
Zheren Fu, Lei Zhang, Hou Xia, and Zhendong Mao.
\newblock Linguistic-aware patch slimming framework for fine-grained cross-modal alignment.
\newblock In \emph{Proc. IEEE/CVF Conf. Comput. Vis. Pattern Recognit. (CVPR)}, pages 26307--26316, 2024.

\bibitem[Gu et~al.(2018)Gu, Cai, Joty, Niu, and Wang]{gu2018look}
Jiuxiang Gu, Jianfei Cai, Shafiq~R Joty, Li Niu, and Gang Wang.
\newblock Look, imagine and match: Improving textual-visual cross-modal retrieval with generative models.
\newblock In \emph{Proc. IEEE/CVF Conf. Comput. Vis. Pattern Recognit. (CVPR)}, pages 7181--7189, 2018.

\bibitem[He et~al.(2022)He, Chen, Xie, Li, Doll{\'a}r, and Girshick]{he2022masked}
Kaiming He, Xinlei Chen, Saining Xie, Yanghao Li, Piotr Doll{\'a}r, and Ross Girshick.
\newblock Masked autoencoders are scalable vision learners.
\newblock In \emph{CVPR}, pages 16000--16009, 2022.

\bibitem[Huang et~al.(2018)Huang, Wu, Song, and Wang]{huang2018learning}
Yan Huang, Qi Wu, Chunfeng Song, and Liang Wang.
\newblock Learning semantic concepts and order for image and sentence matching.
\newblock In \emph{Proc. IEEE/CVF Conf. Comput. Vis. Pattern Recognit. (CVPR)}, pages 6163--6171, 2018.

\bibitem[Kim et~al.(2023)Kim, Kim, and Kwak]{kim2023improving}
Dongwon Kim, Namyup Kim, and Suha Kwak.
\newblock Improving cross-modal retrieval with set of diverse embeddings.
\newblock In \emph{CVPR}, pages 23422--23431, 2023.

\bibitem[Kiros et~al.(2014)Kiros, Salakhutdinov, and Zemel]{kiros2014unifying}
Ryan Kiros, Ruslan Salakhutdinov, and Richard~S Zemel.
\newblock Unifying visual-semantic embeddings with multimodal neural language models.
\newblock \emph{arXiv preprint arXiv:1411.2539}, 2014.

\bibitem[Lee et~al.(2018)Lee, Chen, Hua, Hu, and He]{lee2018stacked}
Kuang-Huei Lee, Xi Chen, Gang Hua, Houdong Hu, and Xiaodong He.
\newblock Stacked cross attention for image-text matching.
\newblock In \emph{Proc. Eur. Conf. Comput. Vis. (ECCV)}, pages 201--216, 2018.

\bibitem[Li et~al.(2019)Li, Zhang, Li, Li, and Fu]{li2019visual}
Kunpeng Li, Yulun Zhang, Kai Li, Yuanyuan Li, and Yun Fu.
\newblock Visual semantic reasoning for image-text matching.
\newblock In \emph{Proc. IEEE Int. Conf. Comput. Vis. (ICCV)}, pages 4654--4662, 2019.

\bibitem[Li et~al.(2022{\natexlab{a}})Li, Zhang, Li, Li, and Fu]{li2022image}
Kunpeng Li, Yulun Zhang, Kai Li, Yuanyuan Li, and Yun Fu.
\newblock Image-text embedding learning via visual and textual semantic reasoning.
\newblock \emph{IEEE transactions on pattern analysis and machine intelligence}, 45\penalty0 (1):\penalty0 641--656, 2022{\natexlab{a}}.

\bibitem[Li et~al.(2022{\natexlab{b}})Li, Guo, Feng, Hwang, and Xue]{li2022multi}
Zheng Li, Caili Guo, Zerun Feng, Jenq-Neng Hwang, and Xijun Xue.
\newblock Multi-view visual semantic embedding.
\newblock In \emph{Proc. Int. Joint Conf. on Artif. Intell.}, 2022{\natexlab{b}}.

\bibitem[Lin et~al.(2014)Lin, Maire, Belongie, Hays, Perona, Ramanan, Doll{\'a}r, and Zitnick]{lin2014microsoft}
Tsung-Yi Lin, Michael Maire, Serge Belongie, James Hays, Pietro Perona, Deva Ramanan, Piotr Doll{\'a}r, and C~Lawrence Zitnick.
\newblock Microsoft coco: Common objects in context.
\newblock In \emph{Proc. Eur. Conf. Comput. Vis. (ECCV)}, pages 740--755, 2014.

\bibitem[Liu et~al.(2020)Liu, Mao, Zhang, Xie, Wang, and Zhang]{liu2020graph}
Chunxiao Liu, Zhendong Mao, Tianzhu Zhang, Hongtao Xie, Bin Wang, and Yongdong Zhang.
\newblock Graph structured network for image-text matching.
\newblock In \emph{Proc. IEEE/CVF Conf. Comput. Vis. Pattern Recognit. (CVPR)}, pages 10921--10930, 2020.

\bibitem[Liu et~al.(2023{\natexlab{a}})Liu, Li, Wu, and Lee]{liu2023visual}
Haotian Liu, Chunyuan Li, Qingyang Wu, and Yong~Jae Lee.
\newblock Visual instruction tuning.
\newblock \emph{Advances in neural information processing systems}, 36:\penalty0 34892--34916, 2023{\natexlab{a}}.

\bibitem[Liu et~al.(2022)Liu, Liu, Wang, and Liu]{liu2022regularizing}
Yang Liu, Hong Liu, Huaqiu Wang, and Mengyuan Liu.
\newblock Regularizing visual semantic embedding with contrastive learning for image-text matching.
\newblock \emph{IEEE Sign. Process. Letters}, 2022.

\bibitem[Liu et~al.(2023{\natexlab{b}})Liu, Liu, Wang, Meng, and Liu]{liu2023bcan}
Yang Liu, Hong Liu, Huaqiu Wang, Fanyang Meng, and Mengyuan Liu.
\newblock Bcan: Bidirectional correct attention network for cross-modal retrieval.
\newblock \emph{IEEE Transactions on Neural Networks and Learning Systems}, 2023{\natexlab{b}}.

\bibitem[Liu et~al.(2025)Liu, Liu, Huang, and Lv]{liu2025asymmetric}
Yang Liu, Mengyuan Liu, shudong Huang, and Jiancheng Lv.
\newblock Asymmetric visual semantic embedding framework for efficient vision-language alignment.
\newblock In \emph{Proceedings of the 39th Annual AAAI Conference on Artificial Intelligence}, 2025.

\bibitem[Liu et~al.(2021)Liu, Lin, Cao, Hu, Wei, Zhang, Lin, and Guo]{liu2021swin}
Ze Liu, Yutong Lin, Yue Cao, Han Hu, Yixuan Wei, Zheng Zhang, Stephen Lin, and Baining Guo.
\newblock Swin transformer: Hierarchical vision transformer using shifted windows.
\newblock In \emph{Proc. IEEE Int. Conf. Comput. Vis. (ICCV)}, pages 10012--10022, 2021.

\bibitem[Loshchilov and Hutter(2019)]{loshchilov2019decoupled}
Ilya Loshchilov and Frank Hutter.
\newblock Decoupled weight decay regularization.
\newblock In \emph{Proc. Int. Conf. on Learn. Representations}, 2019.

\bibitem[Pan et~al.(2023)Pan, Wu, and Zhang]{pan2023fine}
Zhengxin Pan, Fangyu Wu, and Bailing Zhang.
\newblock Fine-grained image-text matching by cross-modal hard aligning network.
\newblock In \emph{Proc. IEEE/CVF Conf. Comput. Vis. Pattern Recognit. (CVPR)}, pages 19275--19284, 2023.

\bibitem[Plummer et~al.(2015)Plummer, Wang, Cervantes, Caicedo, Hockenmaier, and Lazebnik]{plummer2015flickr30k}
Bryan~A Plummer, Liwei Wang, Chris~M Cervantes, Juan~C Caicedo, Julia Hockenmaier, and Svetlana Lazebnik.
\newblock Flickr30k entities: Collecting region-to-phrase correspondences for richer image-to-sentence models.
\newblock In \emph{Proc. IEEE Int. Conf. Comput. Vis. (ICCV)}, pages 2641--2649, 2015.

\bibitem[Qin et~al.(2022)Qin, Peng, Peng, Wang, and Hu]{qin2022deep}
Yang Qin, Dezhong Peng, Xi Peng, Xu Wang, and Peng Hu.
\newblock Deep evidential learning with noisy correspondence for cross-modal retrieval.
\newblock In \emph{Proceedings of the 30th ACM International Conference on Multimedia}, pages 4948--4956, 2022.

\bibitem[Qin et~al.(2023)Qin, Sun, Peng, Zhou, Peng, and Hu]{qin2023cross}
Yang Qin, Yuan Sun, Dezhong Peng, Joey~Tianyi Zhou, Xi Peng, and Peng Hu.
\newblock Cross-modal active complementary learning with self-refining correspondence.
\newblock \emph{arXiv preprint arXiv:2310.17468}, 2023.

\bibitem[Qu et~al.(2020)Qu, Liu, Cao, Nie, and Tian]{qu2020context}
Leigang Qu, Meng Liu, Da Cao, Liqiang Nie, and Qi Tian.
\newblock Context-aware multi-view summarization network for image-text matching.
\newblock In \emph{Proceedings of the 28th ACM International Conference on Multimedia}, pages 1047--1055, 2020.

\bibitem[Qu et~al.(2021)Qu, Liu, Wu, Gao, and Nie]{qu2021dynamic}
Leigang Qu, Meng Liu, Jianlong Wu, Zan Gao, and Liqiang Nie.
\newblock Dynamic modality interaction modeling for image-text retrieval.
\newblock In \emph{Proceedings of the 44th International ACM SIGIR Conference on Research and Development in Information Retrieval}, pages 1104--1113, 2021.

\bibitem[Radford et~al.(2021)Radford, Kim, Hallacy, Ramesh, Goh, Agarwal, Sastry, Askell, Mishkin, Clark, et~al.]{radford2021learning}
Alec Radford, Jong~Wook Kim, Chris Hallacy, Aditya Ramesh, Gabriel Goh, Sandhini Agarwal, Girish Sastry, Amanda Askell, Pamela Mishkin, Jack Clark, et~al.
\newblock Learning transferable visual models from natural language supervision.
\newblock \emph{arXiv preprint arXiv:2103.00020}, 2021.

\bibitem[Song and Soleymani(2019)]{song2019polysemous}
Yale Song and Mohammad Soleymani.
\newblock Polysemous visual-semantic embedding for cross-modal retrieval.
\newblock In \emph{Proc. IEEE/CVF Conf. Comput. Vis. Pattern Recognit. (CVPR)}, pages 1979--1988, 2019.

\bibitem[Wang et~al.(2020)Wang, Zhang, Ji, Pang, and Ma]{wang2020consensus}
Haoran Wang, Ying Zhang, Zhong Ji, Yanwei Pang, and Lin Ma.
\newblock Consensus-aware visual-semantic embedding for image-text matching.
\newblock In \emph{Proc. Eur. Conf. Comput. Vis. (ECCV)}, pages 18--34, 2020.

\bibitem[Wang et~al.(2019)Wang, Liu, Li, Sheng, Yan, Wang, and Shao]{wang2019camp}
Zihao Wang, Xihui Liu, Hongsheng Li, Lu Sheng, Junjie Yan, Xiaogang Wang, and Jing Shao.
\newblock Camp: Cross-modal adaptive message passing for text-image retrieval.
\newblock In \emph{Proc. IEEE Int. Conf. Comput. Vis. (ICCV)}, pages 5764--5773, 2019.

\bibitem[Wei et~al.(2020)Wei, Xu, Yang, Ji, Wang, and Shen]{wei2020universal}
Jiwei Wei, Xing Xu, Yang Yang, Yanli Ji, Zheng Wang, and Heng~Tao Shen.
\newblock Universal weighting metric learning for cross-modal matching.
\newblock In \emph{Proceedings of the IEEE/CVF conference on computer vision and pattern recognition}, pages 13005--13014, 2020.

\bibitem[Zhang et~al.(2022)Zhang, Mao, Wang, and Zhang]{zhang2022negative}
Kun Zhang, Zhendong Mao, Quan Wang, and Yongdong Zhang.
\newblock Negative-aware attention framework for image-text matching.
\newblock In \emph{Proc. IEEE/CVF Conf. Comput. Vis. Pattern Recognit. (CVPR)}, pages 15661--15670, 2022.

\bibitem[Zhang et~al.(2020)Zhang, Lei, Zhang, and Li]{zhang2020context}
Qi Zhang, Zhen Lei, Zhaoxiang Zhang, and Stan~Z Li.
\newblock Context-aware attention network for image-text retrieval.
\newblock In \emph{Proc. IEEE/CVF Conf. Comput. Vis. Pattern Recognit. (CVPR)}, pages 3536--3545, 2020.

\end{thebibliography}
}


\end{document}